\documentclass{ecai2014}
\usepackage{xspace}
\usepackage{color}
\usepackage{amsmath,amssymb,amsthm}
\usepackage{multirow}
\usepackage{tikz}
\usetikzlibrary{arrows,positioning,shapes.misc,shapes.geometric}
\usetikzlibrary{plotmarks}
\usepackage{subfig}
\usepackage{comment}
\usepackage{algorithm}
\usepackage[noend]{algpseudocode}
\usepackage{hyperref}
\usepackage{times}

\newtheorem{proposition}{Proposition}
\newtheorem{corollary}{Corollary}
\theoremstyle{definition}
\newtheorem{definition}{Definition}
\newtheorem{example}{Example}

\newcommand{\atoms}{\ensuremath{\mathsf{At}}\xspace}					
\newcommand{\langProp}[1]{\ensuremath{\mathcal{L}(#1)}\xspace}		
\newcommand{\true}{\ensuremath{\mathsf{true}}}						
\newcommand{\false}{\ensuremath{\mathsf{false}}}						
\newcommand{\interpretationsProp}[1]{\ensuremath{\mathsf{Int}(#1)}}	
\newcommand{\modelSet}[1]{\ensuremath{\mathsf{Mod}(#1)}}			
\newcommand{\kb}{\ensuremath{\mathcal{K}}}				
\newcommand{\stream}{\ensuremath{\mathcal{S}}}				
\newcommand{\inc}{\ensuremath{\mathcal{I}}}		
\newcommand{\sinc}{\ensuremath{\mathcal{J}}}		
\newcommand{\allkbs}{\ensuremath{\mathbb{K}}}		
\newcommand{\allstreams}{\ensuremath{\mathbb{S}}}		
\usepackage{float}
\restylefloat{table}
\newcommand{\Free}{\ensuremath{\mathsf{Free}}}
\newcommand{\MI}{\ensuremath{\textsf{MI}}}

\title{Towards Large-scale Inconsistency Measurement\footnote{This paper has also been published in the Proceedings of the 37th German Conference on Artificial Intelligence (KI 2014)}}
\author{Matthias Thimm\institute{Institute for Web Science and Technologies, University of Koblenz-Landau, Germany, \texttt{thimm@uni-koblenz.de}}}

\begin{document}

\maketitle

\begin{abstract}
	We investigate the problem of inconsistency measurement on large knowledge bases by considering \emph{stream-based inconsistency measurement}, i.\,e., we investigate inconsistency measures that cannot consider a knowledge base as a whole but process it within a stream. For that, we present, first, a novel inconsistency measure that is apt to be applied to the streaming case and, second, stream-based approximations for the new and some existing inconsistency measures. We conduct an extensive empirical analysis on the behavior of these inconsistency measures on large knowledge bases, in terms of runtime, accuracy, and scalability. We conclude that for two of these measures, the approximation of the new inconsistency measure and an approximation of the \emph{contension} inconsistency measure, large-scale inconsistency measurement is feasible. 
\end{abstract}

\section{Introduction}
Inconsistency measurement \cite{Grant:2006} is a subfield of Knowledge Representation and Reasoning (KR) that is concerned with the quantitative assessment of the severity of inconsistencies in knowledge bases. Consider the following two knowledge bases $\kb_{1}$ and $\kb_{2}$ formalized in propositional logic:
\begin{align*}
	\kb_{1} & = \{a,b\vee c, \neg a \wedge \neg b,d\} & 
	\kb_{2} & = \{a, \neg a, b, \neg b\}
\end{align*}
Both knowledge bases are classically inconsistent as for $\kb_{1}$ we have $\{a,\neg a \wedge \neg b\}\models \perp$ and for $\kb_{2}$ we have, e.\,g., $\{a, \neg a\}\models \perp$. These inconsistencies render the knowledge bases useless for reasoning if one wants to use classical reasoning techniques. In order to make the knowledge bases useful again, one can either use non-monotonic/para\-consistent reasoning techniques \cite{Makinson:2005,Priest:1979} or one revises the knowledge bases appropriately to make them consistent \cite{Hansson:2001}. Looking again at the knowledge bases $\kb_{1}$ and $\kb_{2}$ one can observe that the \emph{severity} of their inconsistency is different. In $\kb_{1}$, only two out of four formulas ($a$ and $\neg a \wedge \neg b$) are \emph{participating} in making $\kb_{1}$ inconsistent while for $\kb_{2}$ all formulas contribute to its inconsistency. Furthermore, for $\kb_{1}$ only two propositions ($a$ and $b$) participate in a conflict and using, e.\,g., paraconsistent reasoning one could still infer meaningful statements about $c$ and $d$. For $\kb_{2}$ no such statement can be made. This leads to the assessment that $\kb_{2}$ should be regarded \emph{more} inconsistent than $\kb_{1}$. Inconsistency measures can be used to quantitatively assess the inconsistency of knowledge bases and to provide a guide for how to repair them, cf. \cite{Grant:2011}. Moreover, they can be used as an analytical tool to assess the quality of knowledge representation. For example, one simple inconsistency measure is to take the number of \emph{minimal inconsistent subsets} (\textsf{MI}s) as an indicator for the inconsistency: the more \textsf{MI}s a knowledge base contains, the more inconsistent it is. For $\kb_{1}$ we have then $1$ as its inconsistency value and for $\kb_{2}$ we have $2$. 

In this paper, we consider the computational problems of inconsistency measurement, particularly with respect to scalable inconsistency measurement on large knowledge bases, as they appear in, e.\,g., Semantic Web applications. To this end we present a novel inconsistency measure $\inc_{hs}$ that approximates the $\eta$-inconsistency measure from \cite{Knight:2002} and is particularly apt to be applied to large knowledge bases. This measure is based on the notion of a \emph{hitting set} which (in our context) is a minimal set of classical interpretations such that every formula of a knowledge base is satisfied by at least one element of the set. In order to investigate the problem of measuring inconsistency in large knowledge bases we also present a stream-based processing framework for inconsistency measurement. More precisely, the contributions of this paper are as follows:
\begin{enumerate}
	\item We present a novel inconsistency measure $\inc_{hs}$ based on hitting sets and show how this measure relates to other measures and, in particular, that it is a simplification of the $\eta$-inconsistency measure \cite{Knight:2002} (Section~\ref{sec:hsmeasure}).
	\item We formalize a theory of inconsistency measurement in streams and provide approximations of several inconsistency measures for the streaming case (Section~\ref{sec:streams}).
	\item We conduct an extensive empirical study on the behavior of those inconsistency measures in terms of runtime, accuracy, and scalability. In particular, we show that the stream variants of $\inc_{hs}$ and of the \emph{contension} measure $\inc_{c}$ are effective and accurate for measuring inconsistency in the streaming setting and, therefore, in large knowledge bases (Section~\ref{sec:eval}).
\end{enumerate}
We give necessary preliminaries for propositional logic and inconsistency measurement in Section~\ref{sec:preliminaries} and conclude the paper with a discussion in Section~\ref{sec:discussion}. Proofs of technical results can be found in Appendix~\ref{app:proofs}.

\section{Preliminaries}\label{sec:preliminaries}
Let $\atoms$ be a propositional signature, i.\,e., a (finite) set of propositions, and let $\langProp{\atoms}$ be the corresponding propositional language, constructed using the usual connectives $\wedge$ (\emph{and}), $\vee$ (\emph{or}), and $\neg$ (\emph{negation}).
We use the symbol $\perp$ to denote contradiction.
	Then a knowledge base $\kb$ is a finite set of formulas $\kb\subseteq\langProp{\atoms}$. Let $\allkbs(\atoms)$ be the set of all knowledge bases.
We write $\allkbs$ instead of $\allkbs(\atoms)$ when there is no ambiguity regarding the signature.
Semantics to $\langProp{\atoms}$ is given by \emph{interpretations} $\omega:\atoms\rightarrow\{\true,\false\}$.
Let $\interpretationsProp{\atoms}$ denote the set of all interpretations for \atoms. An interpretation $\omega$ \emph{satisfies} (or is a \emph{model} of) an atom $a\in\atoms$, denoted by $\omega\models a$ (or $\omega\in\modelSet{a})$, if and only if $\omega(a)=\true$. Both $\models$ and $\modelSet{\cdot}$ are extended to arbitrary formulas, sets, and knowledge bases as usual.

Inconsistency measures are functions $\inc:\allkbs\rightarrow[0,\infty)$ that aim at assessing the severity of the inconsistency in a knowledge base $\kb$, cf. \cite{Grant:2011}. The basic idea is that the larger the inconsistency in $\kb$ the larger the value $\inc(\kb)$. However, inconsistency is a concept that is not easily quantified and there have been a couple of proposals for inconsistency measures so far, see e.\,g. \cite{Knight:2002,Ma:2009,Grant:2013,Grant:2006,Hunter:2010,Thimm:2013}. There are two main paradigms for assessing inconsistency \cite{Hunter:2010}, the first being based on the (number of) formulas needed to produce inconsistencies and the second being based on the proportion of the language that is affected by the inconsistency. Below we recall some popular measures from both categories but we first introduce some necessary notations. 
Let $\kb\in\allkbs$ be some knowledge base. 
\begin{definition}
	A set $M\subseteq\kb$ is called \emph{minimal inconsistent subset} (\textsf{MI}) of $\kb$ if $M\models \perp$ and there is no $M'\subset M$ with $M'\models \perp$. Let $\MI(\kb)$ be the set of all \textsf{MI}s of $\kb$.
\end{definition}
\begin{definition}
	A formula $\alpha\in\kb$ is called \emph{free formula} of $\kb$ if there is no $M\in\MI(\kb)$ with $\alpha\in M$. Let $\Free(\kb)$ denote the set of all free formulas of $\kb$.
\end{definition}
We adopt the following definition of a (basic) inconsistency measure from \cite{Grant:2011}.
\begin{definition}\label{def:inc}
	A \emph{basic inconsistency measure} is a function $\inc:\allkbs\rightarrow[0,\infty)$ that satisfies the following three conditions:
	\begin{enumerate}
		\item $\inc(\kb)=0$ if and only if $\kb$ is consistent,
		\item if $\kb\subseteq\kb'$ then $\inc(\kb)\leq \inc(\kb')$, and
		\item for all $\alpha\in \Free(\kb)$ we have $\inc(\kb)=\inc(\kb\setminus\{\alpha\})$.
	\end{enumerate}
\end{definition}
The first property (also called \emph{consistency}) of a basic inconsistency measure ensures that all consistent knowledge bases receive a minimal inconsistency value and every inconsistent knowledge base receive a positive inconsistency value. The second property (also called \emph{monotony}) states that the value of inconsistency can only increase when adding new information. The third property (also called \emph{free formula independence}) states that removing harmless formulas from a knowledge base---i.\,e., formulas that do not contribute to the inconsistency---does not change the value of inconsistency.
For the remainder of this paper we consider the following selection of inconsistency measures: 
the $\MI$ measure $\inc_{\MI}$, the $\MI^{c}$ measure $\inc_{\MI^{c}}$, the contension measure $\inc_{c}$, and the $\eta$ measure $\inc_{\eta}$, which will be defined below, cf. \cite{Grant:2011,Knight:2002}. In order to define the contension measure $\inc_{c}$ we need to consider three-valued interpretations for propositional logic \cite{Priest:1979}. A three-valued interpretation $\upsilon$ on $\atoms$ is a function $\upsilon:\atoms\rightarrow\{T,F,B\}$ where the values $T$ and $F$ correspond to the classical $\true$ and $\false$, respectively. 
The additional truth value $B$ stands for \emph{both} and is meant to represent a conflicting truth value for a proposition. The function $\upsilon$ is extended to arbitrary formulas as shown in Table~\ref{tbl:para}.
\begin{table*}
	\begin{center}
	\begin{tabular}{cc|c|c|c||cc|c|c|c||cc|c|c|c}
		$\alpha$ 	& $\beta$ 	& $\alpha\wedge\beta$ 	&$\alpha\vee\beta$ 	& $\neg\alpha$ & $\alpha$ 	& $\beta$ 	& $\alpha\wedge\beta$ 	&$\alpha\vee\beta$ 	& $\neg\alpha$ & $\alpha$ 	& $\beta$ 	& $\alpha\wedge\beta$ 	&$\alpha\vee\beta$ 	& $\neg\alpha$\\\hline
		T		   	& 	T		&	T					& T						&	F			& B		   	& 	T		&			B			&	T					&		B		& F		   	& 	T		&			F			&		T				&		T		\\
		T		   	& 	B		&	B					&	T					&	F			& B		   	& 	B		&			B			&	B					&		B		& F		   	& 	B		&			F			&		B				&		T		\\
		T		   	& 	F		&	F					&	T					&	F			& B		   	& 	F		&			F			&	B					&		B		& F		   	& 	F		&			F			&		F				&		T		
	\end{tabular}
	\caption{Truth tables for propositional three-valued logic \cite{Priest:1979}.}
	\label{tbl:para}
	\end{center}
\end{table*}
Then, an interpretation $\upsilon$ satisfies a formula $\alpha$, denoted by $\upsilon\models^{3}\alpha$ if either $\upsilon(\alpha)=T$ or $\upsilon(\alpha)=B$.

For defining the $\eta$-inconsistency measure \cite{Knight:2002} we need to consider probability functions $P$ of the form $P:\interpretationsProp{\atoms}\rightarrow [0,1]$ with $\sum_{\omega\in\interpretationsProp{\atoms}}P(\omega)=1$. Let $\mathcal{P}(\atoms)$ be the set of all those probability functions and for a given probability function $P\in\mathcal{P}(\atoms)$ define the probability of an arbitrary formula $\alpha$ via $P(\alpha)=\sum_{\omega\models\alpha}P(\omega)$.
\begin{definition}
	Let $\inc_{\MI}$, $\inc_{\MI^{c}}$, $\inc_{c}$, and $\inc_{\eta}$ be defined via 
	\begin{align*}
		\inc_{\MI}(\kb)	& = |\MI(\kb)|,\\
		\inc_{\MI^c}(\kb)& = \sum_{M\in\MI(\kb)}\frac{1}{|M|},\\
		\inc_{c}(\kb)	& = \min\{|\upsilon^{-1}(B)|\mid\upsilon\models^{3} \kb\},\\
		\inc_{\eta}(\kb)	& = 1-\max\{\xi\mid\exists P\in \mathcal{P}(\atoms):\forall \alpha\in\kb:P(\alpha)\geq \xi\}
	\end{align*}
\end{definition}
The measure $\inc_{\MI}$ takes the number of minimal inconsistent subsets of a knowledge base as an indicator for the amount of inconsistency: the more minimal inconsistent subsets the more severe the inconsistency. The measure $\inc_{\MI^{c}}$ refines this idea by also taking the size of the minimal inconsistent subsets into account. Here the idea is that larger minimal inconsistent subsets indicate are less severe than smaller minimal inconsistent subsets (the less formulas are needed to produce an inconsistency the more ``obvious'' the inconsistency). The measure $\inc_{c}$ considers the set of three-valued models of a knowledge base (which is always non-empty) and uses the minimal number of propositions with conflicting truth value as an indicator for inconsistency. Finally, the measure $\inc_{\eta}$ (which always assigns an inconsistency value between $0$ and $1$) looks for the maximal probability one can assign to every formula of a knowledge base.
All these measures are basic inconsistency measures as defined in Definition~\ref{def:inc}.
\begin{example}\label{ex:ex1}
	For the knowledge bases $\kb_{1}= \{a,b\vee c, \neg a \wedge \neg b,d\}$ and $\kb_{2}  = \{a, \neg a, b,$ $\neg b\}$ from the introduction we obtain
$\inc_{\MI}(\kb_{1})=1$, $\inc_{\MI^c}(\kb_{1})=0.5$, $\inc_{c}(\kb_{1})=2$, $\inc_{\eta}(\kb_{1})=0.5$, $\inc_{\MI}(\kb_{2})=2$, $\inc_{\MI^c}(\kb_{2})=1$, $\inc_{c}(\kb_{2})=2$, $\inc_{\eta}(\kb_{2})=0.5$.
\end{example}
For a more detailed introduction to inconsistency measures see e.\,g. \cite{Grant:2006,Grant:2011,Knight:2002} and for some recent developments see e.\,g.\ \cite{Grant:2013,Jabbour:2014}.

As for computational complexity, the problem of computing an inconsistency value wrt. any of the above inconsistency measures is at least FNP-hard\footnote{FNP is the generalization of the class NP to functional problems.} as it contains a satisfiability problem as a sub problem.

\section{An Inconsistency Measure based on Hitting Sets}\label{sec:hsmeasure}
The basic idea of our novel inconsistency measure $\inc_{hs}$ is inspired by the measure $\inc_{\eta}$ which seeks a probability function that maximizes the probability of all formulas of a knowledge base. Basically, the measure $\inc_{\eta}$ looks for a minimal number of models of parts of the knowledge base and maximizes their probability in order to maximize the probability of the formulas. By just considering this basic idea we arrive at the notion of a \emph{hitting set} for inconsistent knowledge bases.
\begin{definition}
	A subset $H\subset \interpretationsProp{\atoms}$ is called a \emph{hitting set} of $\kb$ if for every $\alpha\in\kb$ there is $\omega\in H$ with $\omega\models \alpha$. $H$ is called a \textsf{card}-minimal hitting set if it is minimal wrt. cardinality. Let $h_{\kb}$ be the cardinality of any \textsf{card}-minimal hitting set (define $h_{\kb}=\infty$ if there does not exist a hitting set of $\kb$).
\end{definition}
\begin{definition}
	The function $\inc_{hs}:\allkbs\rightarrow[0,\infty]$ is defined via
$
		\inc_{hs}(\kb)  = h_{\kb}-1
$ 
	for every $\kb\in\allkbs$.
\end{definition}
Note, that if a knowledge base $\kb$ contains a contradictory formula (e.\,g. $a\wedge\neg a$) we have $\inc_{hs}(\kb)=\infty$. In the following, we assume that $\kb$ contains no such contradictory formulas.
\begin{example}
	Consider the knowledge base $\kb_{3}$ defined via
	\begin{align*}
		\kb_{3} & = \{a\vee d, a\wedge b \wedge c, b, \neg b \vee \neg a, a\wedge b \wedge \neg c, a\wedge \neg b \wedge c\}
	\end{align*}
	Then $\{\omega_{1},\omega_{2},\omega_{3}\}\subset \interpretationsProp{\atoms}$ with $\omega_{1}(a)=\omega_{1}(b)=\omega_{1}(c)=\true$, $\omega_{2}(a)=\omega_{2}(c)=\true$, $\omega_{1}(b)=\false$, and $\omega_{3}(a)=\omega_{3}(b)=\true$, $\omega_{3}(c)=\false$ is a \textsf{card}-minimal hitting set for $\kb_{3}$ and therefore $\inc_{hs}(\kb_{3})=2$.
	Note that for the knowledge bases $\kb_{1}$ and $\kb_{2}$ from Example~\ref{ex:ex1} we have $\inc_{hs}(\kb_{1})=\inc_{hs}(\kb_{2})=1$.
\end{example}
\begin{proposition}
	The function $\inc_{hs}$ is a (basic) inconsistency measure.
\end{proposition}
The result below shows that $\inc_{hs}$ also behaves well with some more properties mentioned in the literature \cite{Hunter:2010,Thimm:2013}. For that, we denote with $\atoms(F)$ for a formula or a set of formulas $F$ the set of propositions appearing in $F$. Furthermore, two knowledge bases $\kb_{1}$, $\kb_{2}$ are \emph{semi-extensionally equivalent} ($\kb_{1}\equiv^{\sigma}\kb_{2}$) if there is a bijection $\sigma:\kb_{1}\rightarrow\kb_{2}$ such that for all $\alpha\in\kb_{1}$ we have $\alpha\equiv \sigma(\alpha)$. 
\begin{proposition}	
	The measure $\inc_{hs}$ satisfies the following properties:
	\begin{itemize}
		\item If $\alpha\in\kb$ is such that $\atoms(\alpha)\cap\atoms(\kb\setminus\{\alpha\})=\emptyset$ then $\inc_{hs}(\kb)=\inc_{hs}(\kb\setminus\{\alpha\})$ (\emph{safe formula independence}).
		\item If $\kb\equiv^{\sigma}\kb'$ then $\inc_{hs}(\kb)=\inc_{hs}(\kb')$ (\emph{irrelevance of syntax}).
		\item If $\alpha\models\beta$ and $\alpha\not\models \perp$ then $\inc_{hs}(\kb\cup\{\alpha\})\geq \inc_{hs}(\kb\cup\{\beta\})$ (\emph{dominance}).
	\end{itemize}
\end{proposition}
The measure $\inc_{hs}$ can also be nicely characterized by a consistent \emph{partitioning} of a knowledge base.
\begin{definition}
	A set $\Phi=\{\Phi_{1},\ldots,\Phi_{n}\}$ with $\Phi_{1}\cup\ldots\cup\Phi_{n}=\kb$ and $\Phi_{i}\cap\Phi_{j}=\emptyset$ for $i,j=1,\ldots,n$, $i\neq j$, is called a \emph{partitioning} of $\kb$. A partitioning $\Phi=\{\Phi_{1},\ldots,\Phi_{n}\}$ is consistent if $\Phi_{i}\not\models\perp$ for $i=1,\ldots,n$. A consistent partitioning $\Phi$ is called \textsf{card}-minimal if it is minimal wrt. cardinality among all consistent partitionings of $\kb$.
\end{definition}
\begin{proposition}
	A consistent partitioning $\Phi$ is a \textsf{card}-minimal partitioning of $\kb$ if and only if $\inc_{hs}(\kb) = |\Phi|-1$.
\end{proposition}
As $\inc_{hs}$ is inspired by $\inc_{\eta}$ we go on by comparing these two measures.
\begin{proposition}\label{prop:hseta}
	Let $\kb$ be a knowledge base. If $\infty>\inc_{hs}(\kb)>0$ then
	\begin{align*}
		1-\frac{1}{\inc_{hs}(\kb)} < \inc_{\eta}(\kb) \leq 1-\frac{1}{\inc_{hs}(\kb)+1}
	\end{align*}	
\end{proposition}
Note that for $\inc_{hs}(\kb)=0$ we always have $\inc_{\eta}(\kb)=0$ as well, as both are basic inconsistency measures.
\begin{corollary}
	If $\inc_{\eta}(\kb_{1})\leq \inc_{\eta}(\kb_{2})$ then $\inc_{hs}(\kb_{1})\leq \inc_{hs}(\kb_{2})$.
\end{corollary}
However, the measures $\inc_{\eta}$ and $\inc_{hs}$ are not equivalent as the following example shows.
\begin{example}
	Consider the knowledge bases 
		$\kb_{1}  = \{ a, \neg a \}$ and
		$\kb_{2}  = \{ a, b, \neg a\vee \neg b \}$.
	Then we have $\inc_{hs}(\kb_{1}) = \inc_{hs}(\kb_{2}) = 1$ but $\inc_{\eta}(\kb_{1}) = 0.5 > 1/3 =\inc_{\eta}(\kb_{2})$.
\end{example}
It follows that the order among knowledge bases induced by $\inc_{\eta}$ is a refinement of the order induced by $\inc_{hs}$. However, $\inc_{hs}$ is better suited for approximation in large knowledge bases than $\inc_{\eta}$, cf. the following section.

The idea underlying $\inc_{hs}$ is also similar to the contension inconsistency measure $\inc_{c}$. However, these measures are not equivalent as the following example shows.
\begin{example}
	Consider the knowledge bases $\kb_{1}$ and $\kb_{2}$ given as
	\begin{align*}
		\kb_{1} & = \{ a\wedge b \wedge c, \neg a \wedge \neg b \wedge \neg c \} &
		\kb_{2} & = \{ a \wedge b, \neg a \wedge b, a \wedge \neg b \}
	\end{align*}
Then we have $\inc_{hs}(\kb_{1}) = 2 < 3 =\inc_{hs}(\kb_{2})$ but $\inc_{c}(\kb_{1}) = 3 > 2 =\inc_{c}(\kb_{2})$. 
\end{example}

\section{Inconsistency Measurement in Streams}\label{sec:streams}
In the following, we discuss the problem of inconsistency measurement in large knowledge bases. We address this issue by using a stream-based approach of accessing the formulas of a large knowledge base. Formulas of a knowledge base then need to be processed one by one by a stream-based inconsistency measure. The goal of this formalization is to obtain stream-based inconsistency measures that approximate given inconsistency measures when the latter would have been applied to the knowledge base as a whole. We first formalize this setting and, afterwards, provide concrete approaches for some inconsistency measures.

\subsection{Problem Formalization}
We use a very simple formalization of a stream that is sufficient for our needs.
\begin{definition}
	A \emph{propositional stream} $\stream$ is a function $\stream:\mathbb{N}\rightarrow \langProp{\atoms}$. Let $\allstreams$ be the set of all propositional streams.
\end{definition}
A propositional stream models a sequence of propositional formulas. On a wider scope, a propositional stream can also be interpreted as a very general abstraction of the output of a linked open data crawler (such as LDSpider \cite{ldspider}) that crawls knowledge formalized as \textsf{RDF} (\emph{Resource Description Framework}) from the web, enriched, e.\,g.\, with OWL semantics.
We model large knowledge bases by propositional streams that indefinitely repeat the formulas of the knowledge base. For that, we assume for a knowledge base $\kb=\{ \phi_{1},\ldots,\phi_{n}\}$ the existence of a \emph{canonical enumeration} $\kb^{c}=\langle \phi_{1},\ldots,\phi_{n}\rangle$ of the elements of $\kb$. This enumeration can be arbitrary and has no specific meaning other than to enumerate the elements in an unambiguous way.
\begin{definition}
	Let $\kb$ be a knowledge base and $\kb^{c}=\langle \phi_{1},\ldots,\phi_{n}\rangle$ its canonical enumeration. The \emph{\emph{\kb}-stream} $\stream_{\kb}$ is defined as
	$
		\stream_{\kb}(i) = \phi_{(i\,\mathsf{mod}\, n) + 1}
	$
	for all $i\in\mathbb{N}$.
\end{definition}
Given a $\kb$-stream $\mathcal{S}_{\kb}$ and an inconsistency measure $\inc$ we aim at defining a method that processes the elements of $\mathcal{S}_{\kb}$ one by one and approximates $\inc(\kb)$. 
\begin{definition}
	A \emph{stream-based inconsistency measure} $\sinc$ is a function $\sinc:\allstreams\times\mathbb{N}\rightarrow[0,\infty)$.
\end{definition}
\begin{definition}
	Let $\inc$ be an inconsistency measure and $\sinc$ a stream-based inconsistency measure. Then $\sinc$ \emph{approximates} (or \emph{is an approximation of}) $\inc$ if for all $\kb\in\allkbs$ we have
	$
		\lim_{i\rightarrow\infty}\sinc(\mathcal{S}_{\kb},i)=\inc(\kb)
	$.
\end{definition}
\subsection{A Naive Window-based Approach}
The simplest form of implementing a stream-based variant of any algorithm or function is to use a window-based approach, i.\,e., to consider at any time point a specific excerpt from the stream and apply the original algorithm or function on this excerpt. 
For any propositional stream $\mathcal{S}$ let $\mathcal{S}^{i,j}$ (for $i\leq j$) be the knowledge base obtained by taking the formulas from $\mathcal{S}$ between positions $i$ and $j$, i.\,e., $\mathcal{S}^{i,j}=\{\mathcal{S}(i),\ldots,\mathcal{S}(j)\}$.
\begin{definition}
	Let $\inc$ be an inconsistency measure, $w\in\mathbb{N}\cup\{\infty\}$, and $g$ some function $g:[0,\infty)\times[0,\infty)\rightarrow [0,\infty)$ with $g(x,y)\in[\min\{x,y\},\max\{x,y\}]$. We define the \emph{naive window-based measure} $\sinc^{w,g}_{\inc}:\allstreams\times\mathbb{N}\rightarrow[0,\infty)$ via	
	\begin{align*}
		\sinc^{w,g}_{\inc}(\mathcal{S},i) & = \left\{\begin{array}{ll}
														0	& \text{if~}i=0\\
														g(\inc(\mathcal{S}^{\max\{0,i-w\},i}),\sinc^{w,g}_{\inc}(\mathcal{S},i-1)) & \text{otherwise}
												\end{array}\right.
	\end{align*}
	for every $\mathcal{S}$ and $i\in\mathbb{N}$.
\end{definition}
The function $g$ in the above definition is supposed to be an aggregation function that combines the new obtained inconsistency value $\inc(\mathcal{S}_{\kb}^{\max\{0,i-w\},i})$ with the previous value $\sinc^{w,g}_{\inc}(\mathcal{S},i-1)$. This function can be ,e.\,g., the maximum function $\max$ or a smoothing function $g_{\alpha}(x,y)=\alpha x+(1-\alpha)y$ for some $\alpha\in[0,1]$ (for every $x,y\in[0,\infty)$).

\begin{proposition}
	Let $\inc$ be an inconsistency measure, $w\in\mathbb{N}\cup\{\infty\}$, and $g$ some function $g:[0,\infty)\times[0,\infty)\rightarrow [0,\infty)$ with $g(x,y)\in[\min\{x,y\},\max\{x,y\}]$.
	\vspace*{-2mm}\begin{enumerate}
		\item If $w$ is finite then $\sinc^{w,g}_{\inc}$ is \emph{not} an approximation of $\inc$.
		\item If $w=\infty$ and $g(x,y)>\min\{x,y\}$ if $x\neq y$ then $\sinc^{w,g}_{\inc}$ is an approximation of $\inc$.		
		\item $\sinc^{w,g}_{\inc}(\mathcal{S}_{\kb},i)\leq \inc(\kb)$ for every $\kb\in\allkbs$ and $i\in\mathbb{N}$.
	\end{enumerate}
\end{proposition}
\subsection{Approximation Algorithms for $\inc_{hs}$ and $\inc_{c}$}
The approximation algorithms for $\inc_{hs}$ and $\inc_{c}$ that are presented in this subsection are using concepts of the programming paradigms of \emph{simulated annealing} and \emph{genetic programming} \cite{Lawrence:1987}. Both algorithms follow the same idea and we will only formalize the one for $\inc_{hs}$ and give some hints on how to adapt it for $\inc_{c}$.

The basic idea for the stream-based approximation of $\inc_{hs}$ is as follows. At any processing step we maintain a candidate set $C\in 2^{\interpretationsProp{\atoms}}$ (initialized with the empty set) that approximates a hitting set of the underlying knowledge base. At the beginning of a processing step we make a random choice (with decreasing probability the more formulas we already encountered) whether to remove some element of $C$. This action ensures that $C$ does not contain superfluous elements. Afterwards we check whether there is still an interpretation in $C$ that satisfies the currently encountered formula. If this is not the case we add some random model of the formula to $C$. Finally, we update the previously computed inconsistency value with $|C|-1$, taking also some aggregation function $g$ (as for the naive window-based approach) into account. In order to increase the probability of successfully finding a minimal hitting set we do not maintain a single candidate set $C$ but a (multi-)set $Cand=\{C_{1},\ldots,C_{m}\}$ for some previously specified parameter $m\in\mathbb{N}$ and use the average size of these candidate hitting sets.
\begin{definition}\label{def:streamhs}
	Let $m\in\mathbb{N}$, $g$ some function $g:[0,\infty)\times[0,\infty)\rightarrow [0,\infty)$ with $g(x,y)\in[\min\{x,y\},\max\{x,y\}]$, and $f:\mathbb{N}\rightarrow[0,1]$ some monotonically decreasing function with $\lim_{n\rightarrow\infty}f(n)=0$. We define $\sinc_{hs}^{m,g,f}$ via
	{\abovedisplayskip1mm
	\belowdisplayskip1mm
	\begin{align*}
		\sinc_{hs}^{m,g,f}(\mathcal{S},i) & = \left\{\begin{array}{ll}
														0	& \text{if~}i=0\\
														\mathtt{update}^{m,g,f}_{hs}(\mathcal{S}(i)) & \text{otherwise}
												\end{array}\right.
	\end{align*}}
	for every $\mathcal{S}$ and $i\in\mathbb{N}$. The function $\mathtt{update}^{m,g,f}_{hs}$ is depicted in Algorithm~\ref{algo1}.
\end{definition}
\begin{algorithm}[t]
	\begin{algorithmic}[1]
		\State Initialize $currentValue$ and $Cand$
		\State $N = N+1$
		\State $newValue = 0$
		\ForAll{$C\in Cand$}
			\State $rand$ $\in [0,1]$
			\If{$rand < f(N)$}
				\State Remove some random $\omega$ from $C$
			\EndIf
			\If{$\neg \exists\omega\in C: \omega\models form$}
				\State Add random $\omega\in\modelSet{form}$ to $C$
			\EndIf
			\State $newValue = newValue + (|C|-1)/|Cand|$
		\EndFor
		\State $currentValue = g(newValue,currentValue)$
		\State \Return $currentValue$
	\end{algorithmic}
	\caption{$\mathtt{update}^{m,g,f}_{hs}(form)$}
	\label{algo1}
\end{algorithm}
At the first call of the algorithm $\mathtt{update}^{m,g,f}_{hs}$ the value of $currentValue$ (which contains the currently estimated inconsistency value) is initialized to $0$ and the (mulit-)set $Cand\subseteq 2^{\interpretationsProp{\atoms}}$ (which contains a population of candidate hitting sets) is initialized with $m$ empty sets. The function $f$ can be any monotonically decreasing function with $\lim_{n\rightarrow\infty}f(n)=0$ (this ensures that at any candidate $C$ reaches some stable result). The parameter $m$ increases the probability that at least one of the candidate hitting sets attains the global optimum of a \textsf{card}-minimal hitting set.

As $\sinc_{hs}^{m,g,f}$ is a random process we cannot show that $\sinc_{hs}^{m,g,f}$ is an approximation of $\inc_{hs}$ in the general case. However, we can give the following result.
\begin{proposition}
	For every probability $p\in[0,1)$, $g$ some function $g:[0,\infty)\times[0,\infty)\rightarrow [0,\infty)$ with $g(x,y)\in[\min\{x,y\},\max\{x,y\}]$ and $g(x,y)>\min\{x,y\}$ if $x\neq y$, a monotonically decreasing function $f:\mathbb{N}\rightarrow[0,1]$ with $\lim_{n\rightarrow\infty}f(n)=0$, and $\kb\in\allkbs$ there is $m\in\mathbb{N}$ such that with probability greater or equal $p$ it is the case that
	\begin{align*}
		\lim_{i\rightarrow\infty}\sinc_{hs}^{m,g,f}(\mathcal{S}_{\kb},i)=\inc_{hs}(\kb)
	\end{align*}
\end{proposition}
This result states that $\sinc_{hs}^{m,g,f}$ indeed approximates $\inc_{hs}$ if we choose the number of populations large enough. In the next section we will provide some empirical evidence that even for small values of $m$ results are satisfactory.

Both Definition~\ref{def:streamhs} and Algorithm~\ref{algo1} can be modified slightly in order to approximate $\inc_{c}$ instead of $\inc_{hs}$, yielding a new measure $\sinc_{c}^{m,g,f}$. For that, the set of candidates $Cand$ contains three-valued interpretations instead of sets of classical interpretations. In line~7, we do not remove an interpretation from $C$ but flip some arbitrary proposition from $B$ to $T$ or $F$. Similarly, in line~9 we do not add an interpretation but flip some propositions to $B$ in order to satisfy the new formula. Finally, the inconsistency value is determined by taking the number of $B$-valued propositions. For more details see the implementations of both $\sinc_{hs}^{m,g,f}$ and $\sinc_{c}^{m,g,f}$, which will also be discussed in the next section.


\section{Empirical Evaluation}\label{sec:eval}
In this section we describe our empirical experiments on runtime, accuracy, and scalability of some stream-based inconsistency measures. Our \textsf{Java} implementations\footnote{
$\inc_{\MI}$, $\inc_{\MI^c}$, $\inc_{\eta}$, $\sinc^{w,g}_{\inc}$:\\ \url{http://mthimm.de/r?r=tweety-inc-commons}\\ 
$\inc_{c}$, $\inc_{hs}$: \url{http://mthimm.de/r?r=tweety-inc-pl}\\
$\sinc_{hs}^{m,g,f}$: \url{http://mthimm.de/r?r=tweety-stream-hs}\\
$\sinc_{c}^{m,g,f}$: \url{http://mthimm.de/r?r=tweety-stream-c}\\
Evaluation framework: \url{http://mthimm.de/r?r=tweety-stream-eval}}
have been added to the \emph{Tweety Libraries for Knowledge Representation} \cite{Thimm:2014}.

\subsection{Evaluated Approaches}
For our evaluation, we considered the inconsistency measures $\inc_{\MI}$, $\inc_{\MI^{c}}$, $\inc_{\eta}$, $\inc_{c}$, and $\inc_{hs}$. We used the SAT solver \emph{lingeling}\footnote{\url{http://fmv.jku.at/lingeling/}} for the sub-problems of determining consistency and to compute a model of a formula. For enumerating the set of \textsf{MI}s of a knowledge base (as required by $\inc_{\MI}$ and $\inc_{\MI^{c}}$) we used MARCO\footnote{\url{http://sun.iwu.edu/~mliffito/marco/}}. The measure $\inc_{\eta}$ was implemented using the linear optimization solver \emph{lp$\_$solve}\footnote{\url{http://lpsolve.sourceforge.net}}.
The measures $\inc_{\MI}$, $\inc_{\MI^{c}}$, and $\inc_{\eta}$ were used to define three different versions of the naive window-based measure $\sinc^{w,g}_{\inc}$ (with $w=500,1000,2000$ and $g=\max$). 
For the measures $\inc_{c}$ and $\inc_{hs}$ we tested each three versions of their streaming variants $\sinc_{c}^{m,g_{0.75},f_{1}}$ and $\sinc_{hs}^{m,g_{0.75},f_{1}}$ (with $m=10,100,500$) with $f_{1}:\mathbb{N}\rightarrow[0,1]$  defined via $f_{1}(i)=1/(i+1)$ for all $i\in\mathbb{N}$ and $g_{0.75}$ is the smoothing function for $\alpha=0.75$ as defined in the previous section. 
\begin{table*}
	\begin{center}\def\arraystretch{1.4}
	\begin{tabular}{c|c|c||c|c|c}
		Measure & RT (iteration)  & RT (total) & 		Measure & RT (iteration)  & RT (total)\\\hline
		$\sinc_{\inc_{\MI}}^{500,\max}$&198ms&133m	&	$\sinc_{c}^{10,g_{0.75},f_{1}}$&0.16ms&6.406s\\
		$\sinc_{\inc_{\MI}}^{1000,\max}$ &359ms&240m&		$\sinc_{c}^{100,g_{0.75},f_{1}}$&1.1ms&43.632s\\
		$\sinc_{\inc_{\MI}}^{2000,\max}$&14703ms&9812m &		$\sinc_{c}^{500,g_{0.75},f_{1}}$&5.21ms&208.422s\\\hline
		$\sinc_{\inc_{\MI^{c}}}^{500,\max}$&198ms&134m&		$\sinc_{hs}^{10,g_{0.75},f_{1}}$&0.07ms&2.788s\\
		$\sinc_{\inc_{\MI^{c}}}^{1000,\max}$&361ms&241m&		$\sinc_{hs}^{100,g_{0.75},f_{1}}$&0.24ms&9.679s\\
		$\sinc_{\inc_{\MI^{c}}}^{2000,\max}$&14812ms&9874m&		$\sinc_{hs}^{500,g_{0.75},f_{1}}$&1.02ms&40.614s
	\end{tabular}
	\end{center}
	\caption{Runtimes for the evaluated measures; each value is averaged over 100 random knowledge bases of 5000 formulas; the total runtime is after 40000 iterations}\label{tbl:runtimes}
\end{table*}

\subsection{Experiment Setup}
For measuring the runtime of the different approaches we generated 100 random knowledge bases in \textsf{CNF} (\emph{Conjunctive Normal Form}) with each 5000 formulas (=disjunctions) and 30 propositions.
For each generated knowledge base $\kb$ we considered its $\kb$-stream and processing of the stream was aborted after 40000 iterations.
We fed the $\kb$-stream to each of the evaluated stream-based inconsistency measures and measured the average runtime per iteration and the total runtime. For each iteration, we set a time-out of 2 minutes and aborted processing of the stream completely if a time-out occurred. 

In order to measure accuracy, for each of the considered approaches we generated another 100 random knowledge bases with specifically set inconsistency values\footnote{The sampling algorithms can be found at\\ \url{http://mthimm.de/r?r=tweety-sampler}}, used otherwise the same settings as above, and measured the returned inconsistency values.

To evaluate the scalability of our stream-based approach of $\inc_{hs}$ we conducted a third experiment\footnote{We did the same experiment with our stream-based approach of $\inc_{c}$ but do not report the results due to the similarity to $\inc_{hs}$ and space restrictions.} where we fixed the number of propositions (60) and the specifically set inconsistency value (200) and varied the size of the knowledge bases from 5000 to 50000 (with steps of 5000 formulas). We measured the total runtime up to the point when the inconsistency value was within a tolerance of $\pm 1$ of the expected inconsistency value.

The experiments were conducted on a server with two Intel Xeon X5550 QuadCore (2.67 GHz) processors with 8 GB RAM running SUSE Linux 2.6.

\subsection{Results}
Our first observation concerns the inconsistency measure $\inc_{\eta}$ which proved to be not suitable to work on large knowledge bases\footnote{More precisely, our implementation of the measure proved to be not suitable for this setting}. Computing the value $\inc_{\eta}(\kb)$ for some knowledge base $\kb$ includes solving a linear optimization problem over a number of variables which is (in the worst-case) exponential in the number of propositions of the signature. In our setting with $|\atoms|=30$ the generated optimization problem contained therefore $2^{30}=1073741824$ variables. Hence, even the optimization problem itself could not be constructed within the timeout of 2 minutes for every step. As we are not aware of any more efficient implementation of $\inc_{\eta}$, we will not report on further results for $\inc_{\eta}$ in the following.

As for the runtime of the naive window-based approaches of $\inc_{\MI}$ and $\inc_{\MI^{c}}$ and our stream-based approaches for $\inc_{c}$ and $\inc_{hs}$ see Table~\ref{tbl:runtimes}. There one can see that $\sinc_{\inc_{\MI}}^{w,g}$ and $\sinc_{\inc_{\MI^{c}}}^{w,g}$ on the one hand, and $\sinc_{c}^{m,g,f}$ and $\sinc_{hs}^{m,g,f}$ on the other hand, have comparable runtimes, respectively. The former two have almost identical runtimes, which is obvious as the determination of the \textsf{MI}s is the main problem in both their computations. Clearly, $\sinc_{c}^{m,g,f}$ and $\sinc_{hs}^{m,g,f}$ are significantly faster per iteration (and in total) than $\sinc_{\inc_{\MI}}^{w,g}$ and $\sinc_{\inc_{\MI^{c}}}^{w,g}$, only very few milliseconds for the latter and several hundreds and thousands of milliseconds for the former (for all variants of $m$ and $w$). The impact of increasing $w$ for $\sinc_{c}^{m,g,f}$ and $\sinc_{hs}^{m,g,f}$ is expectedly linear while the impact of increasing the window size $w$ for $\sinc_{\inc_{\MI}}^{w,g}$ and $\sinc_{\inc_{\MI^{c}}}^{w,g}$ is exponential (this is also clear as both solve an \textsf{FNP}-hard problem). 

\begin{figure}
	\begin{center}
		\subfloat[Accuracy $\sinc_{hs}^{m,g_{0.75},f_{1}}$\label{fig:eval:a}]{
		\begin{tikzpicture}[y=0.3cm, x=.00105cm,font=\sffamily]
			\draw[->] (0,0) -- coordinate (x axis mid) (4100,0);
		    	\draw[->] (0,0) -- coordinate (y axis mid) (0,6.5);
		    	\foreach \x in {1000,2000,3000,4000}
     				\draw (\x,1pt) -- (\x,-3pt)
					node[anchor=north] {\x 0};
		    	\foreach \y in {1,2,3,4,5,6}
     				\draw (1pt,\y) -- (-3pt,\y) 
		     			node[anchor=east] {\y}; 
			\node[below=0.6cm] at (x axis mid) {\#iterations};
			\node[rotate=90,yshift=1cm] at (y axis mid) {Inconsistency value};
		
			\draw[color=gray] plot
				file {data/mes_inc_stream-hs1.txt};
    			\draw[color=lightgray] plot
				file {data/mes_inc_stream-hs2.txt};
			\draw[color=black] plot
				file {data/mes_inc_stream-hs3.txt};
			\draw[dashed] plot 
				file {data/base_inc_stream-hs1.txt};
			\begin{scope}[shift={(1200,1)}] 
				\draw[color=gray] (0,0) -- 
					plot (0.25,0) -- (0.5,0) 
					node[right]{\scriptsize$m=10$};
				\draw[color=lightgray, yshift=\baselineskip*0.7] (0,0) -- 
					plot (0.25,0) -- (0.5,0)
					node[right]{\scriptsize$m=100$};
				\draw[color=black,yshift=2\baselineskip*0.7] (0,0) -- 
					plot (0.25,0) -- (0.5,0)
					node[right]{\scriptsize$m=500$};
			\end{scope}
		\end{tikzpicture}}\\
		\subfloat[Accuracy $\sinc_{c}^{m,g_{0.75},f_{1}}$\label{fig:eval:b}]{
		\begin{tikzpicture}[y=0.3cm, x=.00105cm,font=\sffamily]
			\draw[->] (0,0) -- coordinate (x axis mid) (4100,0);
		    	\draw[->] (0,0) -- coordinate (y axis mid) (0,6.5);
		    	\foreach \x in {1000,2000,3000,4000}
     				\draw (\x,1pt) -- (\x,-3pt)
					node[anchor=north] {\x 0};
		    	\foreach \y in  {1,2,3,4,5,6}
     				\draw (1pt,\y) -- (-3pt,\y) 
		     			node[anchor=east] {\y}; 
			\node[below=0.6cm] at (x axis mid) {\#iterations};
			\node[rotate=90,yshift=1cm] at (y axis mid) {Inconsistency value};
		
			\draw[color=gray] plot
				file {data/mes_inc_stream-c1.txt};
    			\draw[color=lightgray] plot
				file {data/mes_inc_stream-c2.txt};
			\draw[color=black] plot
				file {data/mes_inc_stream-c3.txt};
			\draw[dashed]  plot 
				file {data/base_inc_stream-hs1.txt};
			\begin{scope}[shift={(1200,1)}] 
				\draw[color=gray] (0,0) -- 
					plot (0.25,0) -- (0.5,0) 
					node[right]{\scriptsize$m=10$};
				\draw[color=lightgray, yshift=\baselineskip*0.7] (0,0) -- 
					plot (0.25,0) -- (0.5,0)
					node[right]{\scriptsize$m=100$};
				\draw[color=black,yshift=2\baselineskip*0.7] (0,0) -- 
					plot (0.25,0) -- (0.5,0)
					node[right]{\scriptsize$m=500$};
			\end{scope}
		\end{tikzpicture}}\\
		\subfloat[Accuracy $\sinc_{\inc_{\MI}}^{w,\max}$\label{fig:eval:c}]{
		\begin{tikzpicture}[y=0.1cm, x=.00105cm,font=\sffamily]
			\draw[->] (0,0) -- coordinate (x axis mid) (4100,0);
		    	\draw[->] (0,0) -- coordinate (y axis mid) (0,21);
		    	\foreach \x in {1000,2000,3000,4000}
     				\draw (\x,1pt) -- (\x,-3pt)
					node[anchor=north] {\x 0};
		    	\foreach \y in {5,10,15,20}
     				\draw (1pt,\y) -- (-3pt,\y) 
		     			node[anchor=east] {\y}; 
			\node[below=0.6cm] at (x axis mid) {\#iterations};
			\node[rotate=90,yshift=1cm] at (y axis mid) {Inconsistency value};
		
			\draw[color=gray]  plot
				file {data/mes_inc_stream-wmi1.txt};
    			\draw[color=lightgray]  plot
				file {data/mes_inc_stream-wmi2.txt};
			\draw[color=black] plot
				file {data/mes_inc_stream-wmi3.txt};
			\draw[dashed]  plot 
				file {data/base_inc_wmi1.txt};
			\begin{scope}[shift={(1200,12.5)}] 
					\draw[color=gray] (0,0) -- 
					plot (0.25,0) -- (0.5,0) 
					node[right]{\scriptsize$w=500$};
				\draw[color=lightgray, yshift=\baselineskip*0.7] (0,0) -- 
					plot (0.25,0) -- (0.5,0)
					node[right]{\scriptsize$w=1000$};
				\draw[color=black,yshift=2\baselineskip*0.7] (0,0) -- 
					plot (0.25,0) -- (0.5,0)
					node[right]{\scriptsize$w=2000$};
			\end{scope}
		\end{tikzpicture}}\\
		\subfloat[Accuracy $\sinc_{\inc_{\MI}^{c}}^{w,\max}$\label{fig:eval:d}]{
		\begin{tikzpicture}[y=0.45cm, x=.00105cm,font=\sffamily]
			\draw[->] (0,0) -- coordinate (x axis mid) (4100,0);
		    	\draw[->] (0,0) -- coordinate (y axis mid) (0,5.5);
		    	\foreach \x in {1000,2000,3000,4000}
     				\draw (\x,1pt) -- (\x,-3pt)
					node[anchor=north] {\x 0};
		    	\foreach \y in {1,2,3,4,5}
     				\draw (1pt,\y) -- (-3pt,\y) 
		     			node[anchor=east] {\y}; 
			\node[below=0.6cm] at (x axis mid) {\#iterations};
			\node[rotate=90,yshift=1cm] at (y axis mid) {Inconsistency value};
		
			\draw[color=gray]  plot
				file {data/mes_inc_stream-wmic1.txt};
    			\draw[color=lightgray] plot
				file {data/mes_inc_stream-wmic2.txt};
			\draw[color=black] plot
				file {data/mes_inc_stream-wmic3.txt};
			\draw[dashed]  plot 
				file {data/base_inc_wmic1.txt};
			\begin{scope}[shift={(1200,3.1)}] 
					\draw[color=gray] (0,0) -- 
					plot (0.25,0) -- (0.5,0) 
					node[right]{\scriptsize$w=500$};
				\draw[color=lightgray, yshift=\baselineskip*0.7] (0,0) -- 
					plot (0.25,0) -- (0.5,0)
					node[right]{\scriptsize$w=1000$};
				\draw[color=black,yshift=2\baselineskip*0.7] (0,0) -- 
					plot (0.25,0) -- (0.5,0)
					node[right]{\scriptsize$w=2000$};
			\end{scope}
		\end{tikzpicture}}\\
		\subfloat[Scalability $\sinc_{hs}^{m,g_{0.75},f_{1}}$\label{fig:eval2}]{
		\begin{tikzpicture}[y=0.0008cm, x=.0001cm,font=\sffamily]
			\draw[->] (0,0) -- coordinate (x axis mid) (51000,0);
		    	\draw[->] (0,0) -- coordinate (y axis mid) (0,2300);
		    	
     			\draw (10000,1pt) -- (10000,-3pt) node[anchor=north] {10k};
			\draw (20000,1pt) -- (20000,-3pt) node[anchor=north] {20k};
			\draw (30000,1pt) -- (30000,-3pt) node[anchor=north] {30k};
			\draw (40000,1pt) -- (40000,-3pt) node[anchor=north] {40k};
			\draw (50000,1pt) -- (50000,-3pt) node[anchor=north] {50k};
			
		    	\foreach \y in {500,1000,1500,2000}
     				\draw (1pt,\y) -- (-3pt,\y) 
		     			node[anchor=east] {\y}; 
			\node[below=0.6cm] at (x axis mid) {$|\kb|$};
			\node[rotate=90,yshift=1.3cm] at (y axis mid) {RT in s (total)};
		
			\draw[color=gray]  plot[mark=*, mark options={fill=white}] 
				file {data/scalability_hs1.txt};
    			\draw[color=lightgray] plot[mark=triangle*, mark options={fill=white}] 
				file {data/scalability_hs2.txt};
			\draw[color=black] plot[mark=triangle*, mark options={fill=white}] 
				file {data/scalability_hs3.txt};
			\begin{scope}[shift={(2000,1000)}] 
					\draw[color=gray] (0,0) -- 
					plot[mark=*, mark options={fill=white}] (0.25,0) -- (0.5,0) 
					node[right]{$m=10$};
				\draw[color=lightgray, yshift=\baselineskip] (0,0) -- 
					plot[mark=triangle*, mark options={fill=white}]  (0.25,0) -- (0.5,0)
					node[right]{$m=100$};
				\draw[color=black,yshift=2\baselineskip] (0,0) -- 
					plot[mark=triangle*, mark options={fill=white}]  (0.25,0) -- (0.5,0)
					node[right]{$m=500$};
			\end{scope}
		\end{tikzpicture}}		
		\caption{(a)--(d): Accuracy performance for the evaluated measures (dashed line is actual inconsistency value); each value is averaged over 100 random knowledge bases of 5000 formulas (30 propositions) with varying inconsistency values; (e): Evaluation of the scalability of $\sinc_{hs}^{m,g_{0.75},f_{1}}$; each value is averaged over 10 random knowledge bases of the given size}
		\label{fig:eval}
	\end{center}
\end{figure}
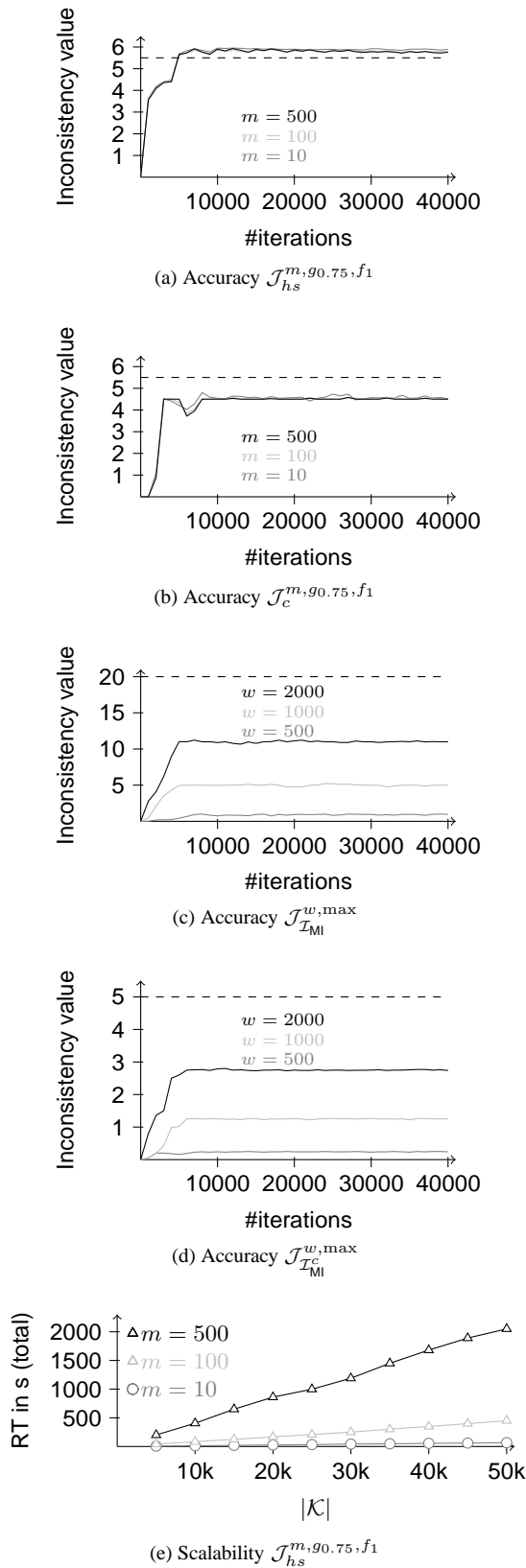

As for the accuracy of the different approaches see Figure~\ref{fig:eval} (a)--(d). There one can see that both $\sinc_{hs}^{m,g,f}$ and $\sinc_{c}^{m,g,f}$ (Figures~\ref{fig:eval:a} and \ref{fig:eval:b}) converge quite quickly (almost right after the knowledge base has been processed once) into a $[-1,1]$ interval around the actual inconsistency value, where $\sinc_{c}^{m,g,f}$ is even closer to it. The naive window-based approaches (Figures~\ref{fig:eval:c} and \ref{fig:eval:d}) have a comparable bad performance (this is clear as those approaches cannot \emph{see} all \textsf{MI}s at any iteration due to the limited window size).
Surprisingly, the impact of larger values of $m$ for $\sinc_{hs}^{m,g,f}$ and $\sinc_{c}^{m,g,f}$ is rather small in terms of accuracy which suggests that the random process of our algorithm is quite robust. Even for $m=10$ the results are quite satisfactory.

As for the scalability of $\sinc_{hs}^{m,g_{0.75},f_{1}}$ see Figure~\ref{fig:eval2}. There one can observe a linear increase in the runtime of all variants wrt. the size of the knowledge base. Furthermore, the difference between the variants is also linearly in the parameter $m$ (which is also clear as each population is an independent random process). It is noteworthy, that the average runtime for $\sinc_{hs}^{10,g_{0.75},f_{1}}$ is about 66.1 seconds for knowledge bases with 50000 formulas. As the significance of the parameter $m$ for the accuracy is also only marginal, the measure $\sinc_{hs}^{10,g_{0.75},f_{1}}$ is clearly an effective and accurate stream-based inconsistency measure.

\section{Discussion and Conclusion}\label{sec:discussion}
In this paper we discussed the issue of large-scale inconsistency measurement and proposed novel approximation algorithms that are effective for the streaming case. To the best of our knowledge, the computational issues for measuring inconsistency, in particular with respect to scalability problems, have not yet been addressed in the literature before. One exception is the work by Ma and colleagues \cite{Ma:2009} who present an anytime algorithm that approximates an inconsistency measure based on a 4-valued paraconsistent logic (similar to the contension inconsistency measure). The algorithm provides lower and upper bounds for this measure and can be stopped at any point in time with some guaranteed quality. The main difference between our framework and the algorithm of \cite{Ma:2009} is that the latter needs to process the whole knowledge base in each atomic step and is therefore not directly applicable for the streaming scenario. The empirical evaluation \cite{Ma:2009} also suggests that our streaming variant of $\inc_{hs}$ is much more performant as Ma et al. report an average runtime of their algorithm of about 240 seconds on a knowledge base with 120 formulas and 20 propositions (no evaluation on larger knowledge bases is given) while our measure has a runtime of only a few seconds for knowledge bases with 5000 formulas with comparable accuracy\footnote{Although hardware specifications for these experiments are different this huge difference is significant.}. A deeper comparison of these different approaches is planned for future work.

Our work showed that inconsistency measurement is not only a theoretical field but can actually be applied to problems of reasonable size. In particular, our stream-based approaches of $\inc_{hs}$ and $\inc_{c}$ are accurate and effective for measuring inconsistencies in large knowledge bases. Current and future work is about the application of our work on linked open data sets \cite{ldspider}.

\bibliographystyle{ecai2014}

\appendix
\section{Proofs of technical results}\label{app:proofs}
\setcounter{proposition}{0}
\setcounter{corollary}{0}
\begin{proposition}
	The function $\inc_{hs}$ is a (basic) inconsistency measure.
	\begin{proof}
		We have to show that properties 1.), 2.), and 3.) of Definition~\ref{def:inc} are satisfied.
		\begin{enumerate}
			\item If $\kb$ is consistent there is a $\omega\in\interpretationsProp{\atoms}$ such that $\omega\models \alpha$ for every $\alpha\in\kb$. Therefore, $H=\{\omega\}$ is a \textsf{card} minimal hitting set and we have $h_{\kb}=1$ and therefore $\inc_{hs}(\kb)=0$. Note that for inconsistent $\kb$ we always have $h_{\kb}>1$.
			\item Let $\kb\subseteq\kb'$ and let $H$ be a \textsf{card}-minimal hitting set of $\kb'$. Then $H$ is also a hitting set of $\kb$ (not necessarily a \textsf{card}-minimal one). Therefore, we have $h_{\kb}\leq h_{\kb'}$ and $\inc_{hs}(\kb)\leq\inc_{hs}(\kb')$.
			\item Let $\alpha\in\Free(\kb)$ and define $\kb'=\kb\setminus\{\alpha\}$.
			Let $H$ be a \textsf{card}-minimal hitting set of $\kb'$ and let $\omega\in H$. Furthermore, let $\kb''\subseteq\kb'$ be the set of all formulas such that $\omega\models\beta$ for all $\beta\in\kb''$. It follows that $\kb''$ is consistent. As $\alpha$ is a free formula it follows that $\kb''\cup \{\alpha\}$ is also consistent (otherwise there would be a minimal inconsistent subset of $\kb''$ containing $\alpha$). Let $\omega'$ be a model of $\kb''\cup \{\alpha\}$. Then $H'=(H\setminus\{\omega\})\cup\{\omega'\}$ is a hitting set of $\kb$ and due to 2.) also \textsf{card}-minimal. Hence, we have $h_{\kb'}=h_{\kb}$ and $\inc_{hs}(\kb')=\inc_{hs}(\kb)$.
		\end{enumerate}
	\end{proof}
\end{proposition}

\begin{proposition}	
	The measure $\inc_{hs}$ satisfies the following properties:
	\begin{itemize}
		\item If $\alpha\in\kb$ is such that $\atoms(\alpha)\cap\atoms(\kb\setminus\{\alpha\})=\emptyset$ then $\inc_{hs}(\kb)=\inc_{hs}(\kb\setminus\{\alpha\})$ (\emph{safe formula independence}).
		\item If $\kb\equiv^{\sigma}\kb'$ then $\inc_{hs}(\kb)=\inc_{hs}(\kb')$ (\emph{irrelevance of syntax}).
		\item If $\alpha\models\beta$ and $\alpha\not\models \perp$ then $\inc_{hs}(\kb\cup\{\alpha\})\geq \inc_{hs}(\kb\cup\{\beta\})$ (\emph{dominance}).
	\end{itemize}
	\begin{proof}~
	\begin{itemize}
		\item This is satisfied as safe formula independence follows from free formula independence, cf. \cite{Hunter:2010,Thimm:2013}.
		\item Let $H$ be a \textsf{card}-minimal hitting set of $\kb$. So, for every $\alpha\in \kb$ we have $\omega\in H$ with $\omega\models \alpha$. Due to $\alpha\equiv\sigma(\alpha)$ we also have $\omega\models\sigma(\alpha)$ and, thus for very $\beta\in\kb'$ we have $\omega\in H$ with $\omega\models\beta$. So $H$ is also a hitting set of $\kb'$. Minimality follows from the fact that $\sigma$ is a bijection. 
		\item Let $H$ be a \textsf{card}-minimal hitting set of $\kb_{1}=\kb\cup\{\alpha\}$ and let $\omega\in H$ be such that $\omega\models\alpha$. Then we also have that $\omega\models \beta$ and $H$ is also a hitting set of $\kb_{2}=\kb\cup\{\beta\}$. Hence, $h_{\kb_{1}}\geq h_{\kb_{2}}$ and $\inc_{hs}(\kb_{1})\geq \inc_{hs}(\kb_{2})$.
	\end{itemize}
	\end{proof}
\end{proposition}

\begin{proposition}
	A consistent partitioning $\Phi$ is a \textsf{card}-minimal partitioning of $\kb$ if and only if $\inc_{hs}(\kb) = |\Phi|-1$.
	\begin{proof}
		Let $\Phi=\{\Phi_{1},\ldots,\Phi_{n}\}$ be a consistent partitioning and let $\omega_{i}\in\interpretationsProp{\atoms}$ be such that $\omega_{i}\models\Phi_{i}$ (for $i=1,\ldots, n$). Then $\{\omega_{1},\ldots,\omega_{n}\}$ is a hitting set of $\kb$ and we have $h_{\kb}\leq |\Phi|$. With the same idea one obtains a consistent partitioning $\Phi$ from every hitting set $H$ of $\kb$ and thus $h_{\kb}\geq |\Phi'|$ for every \textsf{card}-minimal partitioning of $\kb$. Hence, $\inc_{hs}(\kb) = |\Phi|-1$ for every \textsf{card}-minimal partitioning $\Phi$ of $\kb$.
	\end{proof}
\end{proposition}

\begin{proposition}
	Let $\kb$ be a knowledge base. If $\infty>\inc_{hs}(\kb)>0$ then
	\begin{align*}
		1-\frac{1}{\inc_{hs}(\kb)} < \inc_{\eta}(\kb) \leq 1-\frac{1}{\inc_{hs}(\kb)+1}
	\end{align*}	
	\begin{proof}
		For the right inequality, let $H$ be a \textsf{card}-minimal hitting set of $\kb$, i.\,e., we have $\inc_{hs}(\kb)=|H|-1$. Define a probability function $P:\interpretationsProp{\atoms}\rightarrow [0,1]$ via $P(\omega)=1/|H|$ for every $\omega\in H$ and $P(\omega')=0$ for every $\omega'\in \interpretationsProp{\atoms}\setminus H$ (note that $P$ is indeed a probability function). As $H$ is a hitting set of $\kb$ we have that $P(\phi)\geq 1/|H|$ for every $\phi\in\kb$ as at least one model of $\phi$ gets probability $1/|H|$ in $P$. So we have $\inc_{\eta}\leq 1- 1/|H| = 1- 1/(\inc_{hs}(\kb)+1)$.
		For the left inequality we only sketch a proof. Assume that $\inc_{\eta}(\kb)\leq 1/2$, then we have to show that $\inc_{hs}(\kb)<2$ which is equivalent to $\inc_{hs}(\kb)\leq 1$ as the co-domain of $\inc_{hs}$ is a subset of the natural numbers. If $\inc_{\eta}(\kb)\leq 1/2$ then there is a probability function $P$ with $P(\phi)\geq 1/2$ for all $\phi\in \kb$. Let $\Gamma_{P}=\{\omega\in\interpretationsProp{\atoms}\mid P(\omega)>0\}$ and observe $\sum_{\omega\in\Gamma_{P}}P(\omega)=1$. Without loss of generality assume that $P(\omega)=P(\omega')$ for all $\omega,\omega'\in \Gamma_{P}$\footnote{Otherwise let $k\in\mathbb{Q}\cap[0,1]$ be the least common denominator of all $P(\omega)$, $\omega\in\Gamma_{P}$, and replace in $\Gamma_{P}$ every $\omega$ by $k$ duplicates of $\omega$ with probability $P(\omega)/k$ each; for that note that $P$ can always be defined using only rational numbers, cf.\ \cite{Knight:2002}}. Then every $\phi\in \kb$ has to be satisfied by at least half of the interpretations in $\Gamma_{P}$ in order for $P(\phi)=\sum_{\omega\in\Gamma_{P},\omega\models \phi}P(\omega)\geq 1/2$ to hold. Then due to combinatorial reasons there have to be $\omega_{1},\omega_{2}\in \Gamma_{P}$ such that either $\omega_{1}\models\phi$ or $\omega_{2}\models\phi$ for every $\phi\in\kb$. Therefore, $\{\omega_{1},\omega_{2}\}$ is a hitting set and we have $\inc_{hs}(\kb)\leq 1$. By analogous reasoning we obtain $\inc_{hs}(\kb)\leq 2$ if $\inc_{\eta}(\kb)\leq 2/3$ (and therefore $P(\phi)\geq 1/3$ for all $\phi\in\kb$) and the general case $\inc_{hs}(\kb)\leq i$ if $\inc_{\eta}(\kb)\leq (i-1)/i$ and, thus, the claim. Note finally that $\inc_{\eta}(\kb)=1$ if and only if $\kb$ contains a contradictory formula which is equivalent to $\inc_{hs}(\kb)=\infty$ and thus ruled out.
		\end{proof}
\end{proposition}

\begin{corollary}
	If $\inc_{\eta}(\kb_{1})\leq \inc_{\eta}(\kb_{2})$ then $\inc_{hs}(\kb_{1})\leq \inc_{hs}(\kb_{2})$.
	\begin{proof}
		We show the contraposition of the claim, so assume $\inc_{hs}(\kb_{1})> \inc_{hs}(\kb_{2})$ which is equivalent to $\inc_{hs}(\kb_{1})\geq \inc_{hs}(\kb_{2})+1$ as the co-domain of $\inc_{hs}$ is a subset of the natural numbers. By Proposition~\ref{prop:hseta} we have
		\begin{align*}
			\inc_{\eta}(\kb_{1}) & > 1-\frac{1}{\inc_{hs}(\kb_{1})}  \geq 1-\frac{1}{\inc_{hs}(\kb_{2})+1}  \geq \inc_{\eta}(\kb_{2})
		\end{align*}
		which yields $\inc_{\eta}(\kb_{1})>\inc_{\eta}(\kb_{2})$.
	\end{proof}
\end{corollary}

\begin{proposition}
	Let $\inc$ be an inconsistency measure, $w\in\mathbb{N}$, and $g$ some function $g:[0,\infty)\times[0,\infty)\rightarrow [0,\infty)$ with $g(x,y)\in[\min\{x,y\},\max\{x,y\}]$.
	\vspace*{-2mm}\begin{enumerate}
		\item If $w$ is finite then $\sinc^{w,g}_{\inc}$ is \emph{not} an approximation of $\inc$.
		\item If $w=\infty$ and $g(x,y)>\min\{x,y\}$ if $x\neq y$ then $\sinc^{w,g}_{\inc}$ is an approximation of $\inc$.		
		\item $\sinc^{w,g}_{\inc}(\mathcal{S}_{\kb},i)\leq \inc(\kb)$ for every $\kb\in\allkbs$ and $i\in\mathbb{N}$.
	\end{enumerate}
	\begin{proof}~
		\begin{enumerate}
			\item Assume $\kb$ is a minimal inconsistent set with $|\kb|>w$. Then $\inc(\mathcal{S}^{\max\{0,i-w\},i})=0$ for all $i>0$ (as every subset of $\kb$ is consistent) and $\sinc^{w,g}_{\inc}(\mathcal{S},i)=0$ for all $i>0$ as well. As $\inc$ is an inconsistency measure it holds $\inc(\kb)>0$ and, hence, $\sinc^{w,g}_{\inc}$ does not approximate $\inc$.
			\item If $w=\infty$ we have $\inc(\mathcal{S}^{\max\{0,i-w\},i})=\inc(\kb)$ for all $i>i_{0}$ for some $i_{0}\in\mathbb{N}$. As $g(x,y)>\min\{x,y\}$ the value $\inc(\kb)$ will be approximated by $\sinc^{w,g}_{\inc}$ eventually.
			\item This follows from the fact that $\inc$ is a basic inconsistency measure and therefore satisfies $\inc(\kb)\leq \inc(\kb')$ for $\kb\subseteq\kb'$.
		\end{enumerate}	
	\end{proof}
\end{proposition}

\begin{proposition}
	For every probability $p\in[0,1)$, $g$ some function $g:[0,\infty)\times[0,\infty)\rightarrow [0,\infty)$ with $g(x,y)\in[\min\{x,y\},\max\{x,y\}]$ and $g(x,y)>\min\{x,y\}$ if $x\neq y$, a monotonically decreasing function $f:\mathbb{N}\rightarrow[0,1]$ with $\lim_{n\rightarrow\infty}f(n)=0$, and $\kb\in\allkbs$ there is $m\in\mathbb{N}$ such that with probability greater or equal $p$ it is the case that
	$
		\lim_{i\rightarrow\infty}\sinc_{hs}^{m,g,f}(\mathcal{S}_{\kb},i)=\inc_{hs}(\kb)
	$.
	\begin{proof}[Sketch]
		Consider the evolution of single candidate set $C_{1}\in Cand$ during the iterated execution of $\mathtt{update}^{m,g,f}_{hs}(form)$, initialized with the empty set $\emptyset$. Furthermore, let $\hat{C}$ be a \textsf{card}-minimal hitting set of $\kb$. In every iteration the probability of selecting one $\omega\in\hat{C}$ to be added to $C_{1}$ is greater zero as at least one $\omega\in\hat{C}$ is a model of the current formula. Furthermore, the probability of \emph{not} removing any interpretation $\omega'\in C_{1}$ is also greater zero as $f$ is monotonically decreasing (ignoring the very first step). Therefore, the probability $p_{1}$ that $C_{1}$ evolves to $\hat{C}$ (and is not modified thereafter) is greater zero. Furthermore, the evolution of each candidate set $C_{i}\in Cand$ is probabilistically independent of all other evolutions and by considering more candidate sets, i.\,e., by setting the value $m$ large enough, more candidate sets will evolve to some \textsf{card}-minimal hitting set of $\kb$ and the average cardinality of the candidate sets approximates $\inc_{hs}(\kb)+1$.
	\end{proof}
\end{proposition}

\end{document}